\documentclass[a4paper,fleqn]{cas-dc}
\usepackage{multirow}
\usepackage{graphicx}
\usepackage{array}

\usepackage{amsmath}
\usepackage{xcolor}
\usepackage[normalem]{ulem}
\usepackage[numbers]{natbib}

\begin{document}
	\let\WriteBookmarks\relax
	\def\floatpagepagefraction{1}
	\def\textpagefraction{.001}
	\shorttitle{Explainability of RNNs for Enhancing P300-based BCIs}
	\shortauthors{C. Oliva et~al.}
	
	\title  [mode = title]{Explainability of Recurrent Neural Networks for Enhancing P300-based Brain-Computer Interfaces}
	
	\author[1]{C. Oliva}[type=editor,
	auid=000,bioid=1,
	prefix=Dr,
	role=Researcher,
	orcid=0000-0002-8785-6252]
	\cormark[1]
	\ead{christian.oliva@uam.es}
	
	%\credit{Data curation, Writing - Original draft preparation}
	
	\author[2]{V. Changoluisa}[type=editor,
	auid=001,bioid=2,
	prefix=Dr,
	orcid=0000-0002-1255-4954]
	\ead{fchangoluisa@ups.edu.ec}
	
	%\credit{Data curation, Writing - Original draft preparation}
	
	\author[1]{F. B. Rodríguez}[type=editor,
	auid=002,bioid=3,
	prefix=Dr,
	orcid=0000-0003-4053-099X]
	\ead{f.rodriguez@uam.es}
	
	%\credit{Data curation, Writing - Original draft preparation}
	
	\author[1]{L. F. Lago-Fernández}[type=editor,
	auid=003,bioid=4,
	prefix=Dr,
	orcid=0000-0001-8639-8731]
	\ead{luis.lago@uam.es}
	
	%\credit{Data curation, Writing - Original draft preparation}

	\affiliation[1]{organization={Grupo de Neurocomputación Biológica, Departamento de Ingeniería Informática, Escuela Politécnica Superior, Universidad Autónoma de Madrid},
		addressline={c/ Francisco Tomas y Valiente, 11}, 
		city={Madrid},
		%               citysep={}, % Uncomment if no comma needed between city and postcode
		postcode={28049}, 
		state={Madrid},
		country={Spain}}
		
	\affiliation[2]{organization={Grupo de Investigación en Electrónica y Telemática, Universidad Politécnica Salesiana},
		addressline={Campus el Girón, Ave 12 de Octubre 24-22}, 
		city={Quito},
		postcode={170143}, 
		state={Quito},
		country={Ecuador}}

	\cortext[cor1]{Corresponding author}
\begin{abstract}
Brain-Computer Interfaces (BCIs) based on P300 event-related potentials offer promising applications in health, education, and assistive technologies. However, challenges related to inter- and intra-subject variability and the explainability of Deep Learning (DL) models limit their practical deployment. In this work, we present the Post-Recurrent Module (PRM), an additional layer designed to improve both performance and transparency, incorporated into a Recurrent Neural Network (RNN) architecture for classifying P300 signals from EEG data. Our approach enables a dual analysis of spatio-temporal signals through both global and local explainability techniques, allowing us not only to identify the most relevant brain regions and critical time intervals involved in classification, but also to interpret model decisions in terms of spatio-temporal EEG patterns consistent with well-stablished neurophysiological descriptions of the P300. Experimental results show a 9\% improvement in performance over state of the art, while also revealing the importance of inter- and intra-subject variability, in alignment with established neuroscience literature. By making model decisions transparent and efficient, we present a framework for explainable EEG-based models. This framework is not limited to more efficient P300 detection, but can be generalized to a wide range of EEG-based tasks. Its ability to identify key spatial and temporal features makes it suitable for applications such as motor imagery, steady-state visual evoked potentials, and even cognitive workload assessment.
\end{abstract}

\begin{highlights}
	\item A Post-Recurrent Module (PRM) improves RNN performance and interpretability for P300 BCIs
	\item Explainability analysis using gradient-based methods reveals spatio-temporal patterns consistent with established neuroscience findings
	\item Results demonstrate that explainability is essential to understand RNN decision-making, improving reliability
	\item Spatio-temporal explainability enables efficient P300 detection with limited EEG data
	\item Explainability identifies informative electrodes and time windows for P300 decoding
\end{highlights}

\begin{keywords}
	Explainable Artificial Intelligence \sep Electroencephalography \sep Event-Related Potentials \sep Spatio-temporal analysis \sep Post-Recurrent Module \sep Gradient-based Attribution
\end{keywords}

\maketitle

\section{Introduction}
\label{sec:introduction}

Brain-Computer Interfaces (BCIs) are technologies that interpret neuronal activity and convert it into control commands for the management of electronic devices. They enable direct communication between the brain and external systems without relying on muscular activity or peripheral nervous system pathways. This allows the translation of neural signals into actionable outputs for device control. Different areas of knowledge show interest in the study of these technologies since, if they become sufficiently accurate and safe, they could improve people's capabilities in areas such as health, education, or entertainment, among others. One way to implement BCIs is through electroencephalograms (EEGs), which record the brain’s electrical activity non-invasively. After acquiring neurophysiological signals, key features are extracted and translated into control commands that drive external devices or software. These control signals are derived from distinct brain activity patterns, such as event-related potentials (ERPs), steady-state visual-evoked potentials (SSVEPs), or sensorimotor rhythms, each tailored to specific applications like communication, rehabilitation, or assistive technology. Although all these types of brain signals are detected by EEG and utilized in BCIs, they exhibit fundamental differences \cite{cecotti2025,Yi_wei_2019}. In this paper, we focus on a type of ERP called the P300, which is one of the most used ERPs.

The P300 ERP is a positive voltage deflection that occurs approximately 300 ms after presenting a target stimulus. The morphology of this wave is given by its amplitude, latency, and duration time. It has been established as a popular approach to BCIs due to its high temporal resolution, its relative ease of detection, and its association with cognitive processes. However, its application faces important limitations, such as an inherently low signal-to-noise ratio in EEG recordings and a significant dependence on the user's cognitive state, including factors such as age, mental fatigue, motivation, or the intrinsic inter- and intra-subject variability of neural activity \cite{li2020inter}. This variability hinders the generalization capability of data-driven models across sessions and users, thereby limiting their robustness in real-world scenarios and motivating the need for more adaptive and explainable approaches \cite{Yi_wei_2019}.

Despite efforts to understand the exact mechanisms that trigger the P300 wave, the precise nature of the oscillations or brain dynamics that cause its generation and its inter- and intra-subject variability is still unknown \cite{kaplan2005nonstationary}. Thus, differences in amplitude and latency between sessions or individuals raise questions about how genetic, cognitive, and environmental factors shape these responses \cite{kaplan2005nonstationary,zhu2024understanding}. In this regard, several methods have been developed that are specifically tailored to individual user characteristics, which improve the accuracy of P300 detection and mitigate, to some extent, the problem of inter-subject variability \cite{Changoluisa2018,Changoluisa2020,10.1007/978-3-030-85030-2_19}. In this scenario, Machine Learning (ML) has played a fundamental role in the advancement of BCIs. This progress has been made from early algorithms, such as Linear Discriminant Analysis (LDA), or Support Vector Machines (SVM) \cite{lotte2018review}, to more sophisticated algorithms, such as neural networks and Deep Learning (DL) \cite{eeg_oliva21,p300_oliva23_icann,p300_oliva_aiai23}. In the latter case, the non-linear structure of neural networks, in addition to their huge complexity, makes them inherently non-transparent technologies that hide the internal logic. As a result, they are difficult for human experts to interpret and are often regarded as black boxes \cite{guidotti2018survey}, thus hindering the explainability of the obtained results \cite{rajpura2024explainable}. 

Incorporating the study of explainable artificial intelligence (XAI) algorithms used in healthcare systems becomes critical for creating reliable and unbiased systems \cite{wani2024explainable,Wojciech2017}. In the context of P300-based BCIs, it is crucial to advance their applicability and reliability. Since P300 characteristics, such as latency and amplitude, are directly related to complex cognitive processes \cite{Xuejing_2019}, the interpretation of these signals must be transparent and understandable. Interpretable algorithms allow not only the improvement of accuracy in classifying relevant stimuli, but also the identification of individual patterns or anomalies in the signal. In addition, this explainability facilitates the identification of factors that may compromise reliability, such as intra- and inter-subject variability, providing a framework for developing adaptive systems that maintain robust performance even under dynamic conditions.

In this work, we address the implementation of a new Recurrent Neural Network (RNN) architecture for the classification of P300 signals. RNNs are a class of neural networks specifically designed for processing sequential data. Unlike traditional feedforward networks, RNNs incorporate temporal dynamics by maintaining a hidden state that captures information from previous time steps, making them well suited for EEG signal classification. This is particularly relevant for P300 signals, as their temporal structure requires models that can effectively capture dependencies across time steps. Given the challenge of interpretability in DL models, we build upon the previously proposed Post-Recurrent Module (PRM) \cite{p300_oliva23_icann}, which improves the explainability of temporal dependencies by identifying the most relevant temporal samples of the signal for the model's final decision. Furthermore, we apply global and local explainability techniques to analyze the spatial and spatio-temporal characteristics of the model's predictions. The global analysis allows us to understand how different regions of the brain contribute to the classification process, providing a broader perspective on the spatial relevance of input features. Meanwhile, the local analysis explores the spatio-temporal evolution of these contributions, offering a fine-grained explanation of the model's behavior. By combining spatial and spatio-temporal analysis, our approach provides a more comprehensive view of P300 dynamics, an aspect that has received limited attention in the literature. Our findings show that the explainability results align with established knowledge, confirming the validity of our method from both spatial and temporal perspectives.

The remainder of the article is structured as follows. In Section \ref{sec:materials}, we describe the materials and methods used in our work, including the dataset and preprocessing steps, the description of the Recurrent Neural Network architecture, and the experimental design and methodology. Section \ref{sec:rnns} is dedicated to the core of the article, based on the explainability of the RNN model, where we explore its internal behavior, analyze the Post-Recurrent Module, and perform temporal, spatial, and spatio-temporal analyses to shed light on the explainability of the model. Finally, in Section \ref{sec:conclusion}, we discuss our findings, highlight the implications of our approach, and suggest potential directions for further research.

\section{Materials and methods}
\label{sec:materials}

In this section we describe the experimental setup and the methodological framework used in this study. We first introduce the P300-based EEG dataset and detail the preprocessing steps applied to obtain the training data. Then we present the RNN architecture, together with the training configuration and the validation strategy. Finally, we outline the proposed methodology for analyzing the spatio-temporal explainability of RNNs.

\subsection{Dataset Description and Preprocessing}
\label{subsec:dataset}

\begin{figure}
    \centering
    \includegraphics[width=\linewidth]{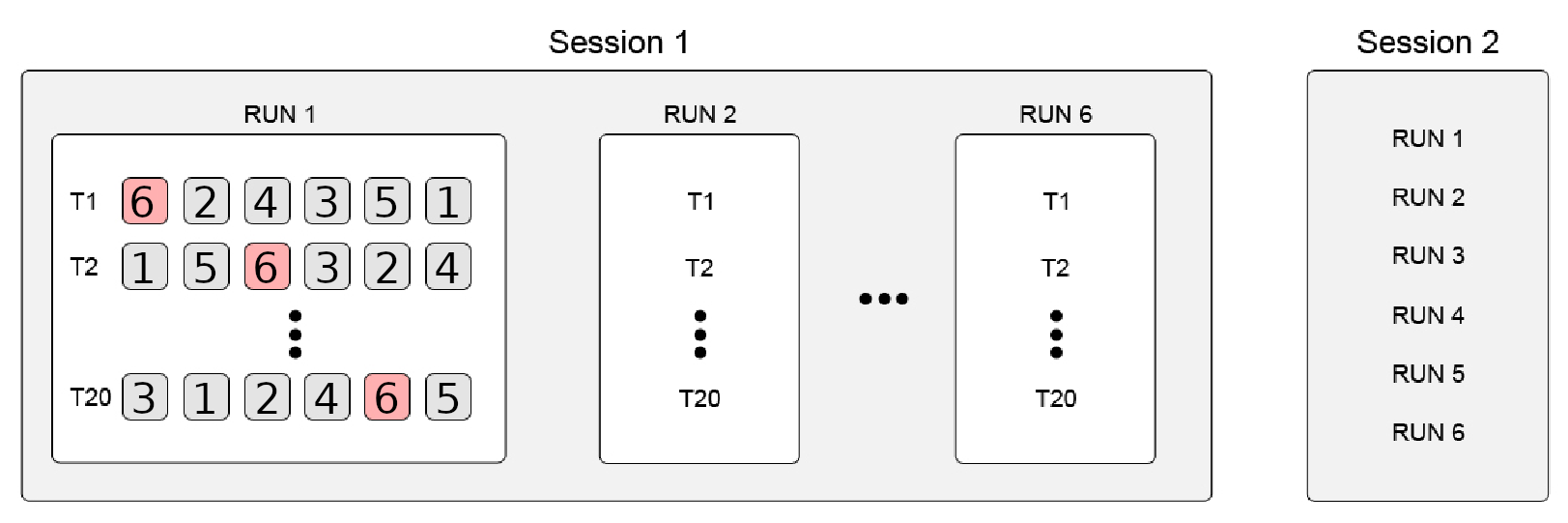}
    \caption{Data description for one day. One day is split into two sessions. Each session contains the subject's EEG signals for six different runs. For each run, the subject's target image is chosen randomly between the six possible images. Each run groups 20 trials, where the six possible images are permuted and flashed to the subject with an ISI of $400 ms.$}
    \label{fig:diagram_sessions}
\end{figure}

We use the dataset defined by Hoffmann et al. \cite{hoffman} based on a six-choice P300 paradigm. The dataset contains electroencephalography (EEG) signals from eight subjects who participated in the experiment. In this experiment, subjects react to sequences of six random images displayed for $100~ms$ with an Inter-Stimulus Interval (ISI) of $400~ms$, where only one of these images is the target stimulus. Thus, there is a gap of $300~ms$ between subsequent images. Each subject participated in this experiment for two days, with each day consisting of two sessions, further divided into six runs. The target stimulus for each run is randomly chosen, and each run comprises 20 trials, in which the images are shuffled and presented to the subject. We illustrate the complete procedure in Figure \ref{fig:diagram_sessions}.

The original dataset recorded the EEG signals at a sampling rate of $2048$ \textit{Hz} from the standard 32 electrodes of the 10-20 international system. The dataset is filtered with a sixth-order forward-backward Butterworth bandpass filter with cut-off frequencies set to $1$ \textit{Hz} and $12$ \textit{Hz}, and its signals are downsampled from $2048$ \textit{Hz} to $32$ \textit{Hz} \cite{hoffman}. This downsampling defines a timestep of $31.25~ms$. 

\begin{figure}
    \centering
    \includegraphics[width=\linewidth]{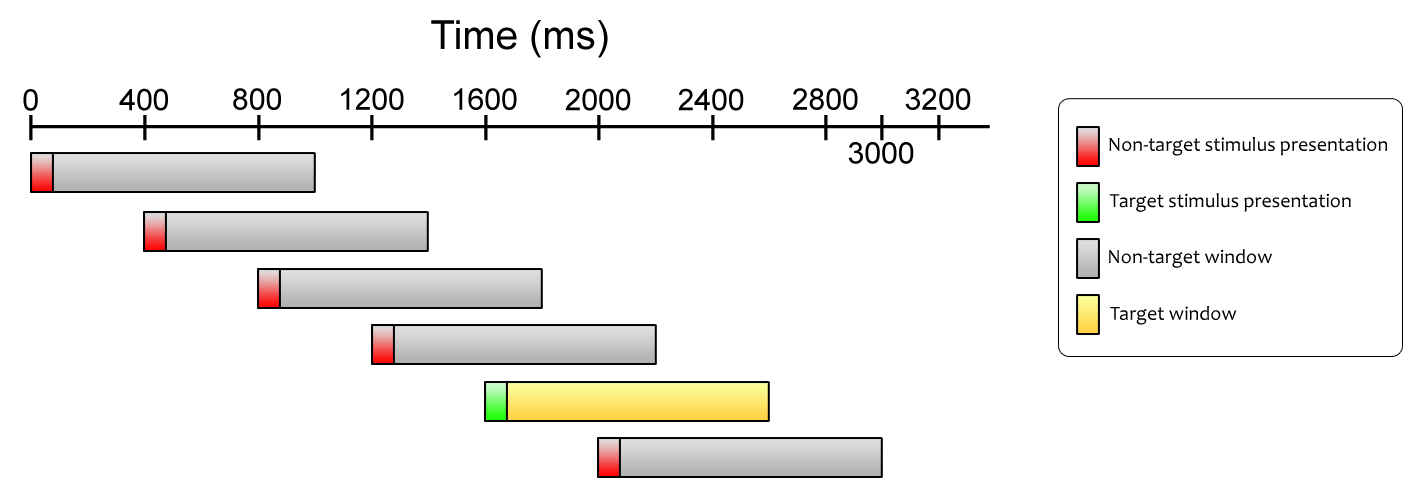}
    \caption{A single trial with the presentation of $6$ images starts at $t = 0~ms$ and finishes at $t = 3000~ms$. Images are presented for $100~ms$ (red/green boxes for non-target/target stimuli, respectively) with an ISI of $400~ms$. The $1000~ms$ windows following each stimulus presentation are used to characterize the user's response to each image.}
    \label{fig:diagram_overlap}
\end{figure}

In oddball paradigms, such as that used in Hoffmann \cite{hoffman}, the P300 typically emerges between $250$-$600~ms$ after stimulus onset, motivating the use of relatively long $1000~ms$ windows to capture the full evolution of the potential. Thus, the standard preprocessing consists of extracting individual $1000~ms$ windows at a sampling rate of $32$ \textit{Hz} (32 timesteps), starting from the onset of the stimuli. Due to the ISI of $400~ms$, this procedure makes the last $600~ms$ of every window overlap with the beginning of the next one. Figure \ref{fig:diagram_overlap} shows an explanatory diagram of this preprocessing for a single trial, where the overlap between contiguous windows is manifest. The classification task consists of predicting the correct label for each window: either \textit{Target stimulus} when the target image is presented to the user (yellow box in the figure) or \textit{Non-Target stimulus} when a non-target image is shown (gray in the figure). The images are flashed during the first $100~ms$ of each window (green and red colors for target and non-target, respectively).

\subsection{Recurrent Neural Network Architecture}
\label{subsec:rnns}

Recurrent Neural Networks (RNNs) are a type of neural network designed to process sequential or time-series data. This is achieved through the addition of recurrent connections, which enable them to learn temporal dependencies within the data. This capability makes RNNs well-suited for many tasks, such as language modeling, speech recognition, and time-series prediction \cite{survey_2,survey_1}. In the case of EEG signals, this temporal modeling is particularly important, since brain activity is not composed of independent values but rather evolves dynamically over time. In this work, we use Elman RNNs \cite{elman_1990}, which are one of the simplest neural network models where recurrent connections are introduced. 

\begin{figure}
    \centering
    \includegraphics[width=\linewidth]{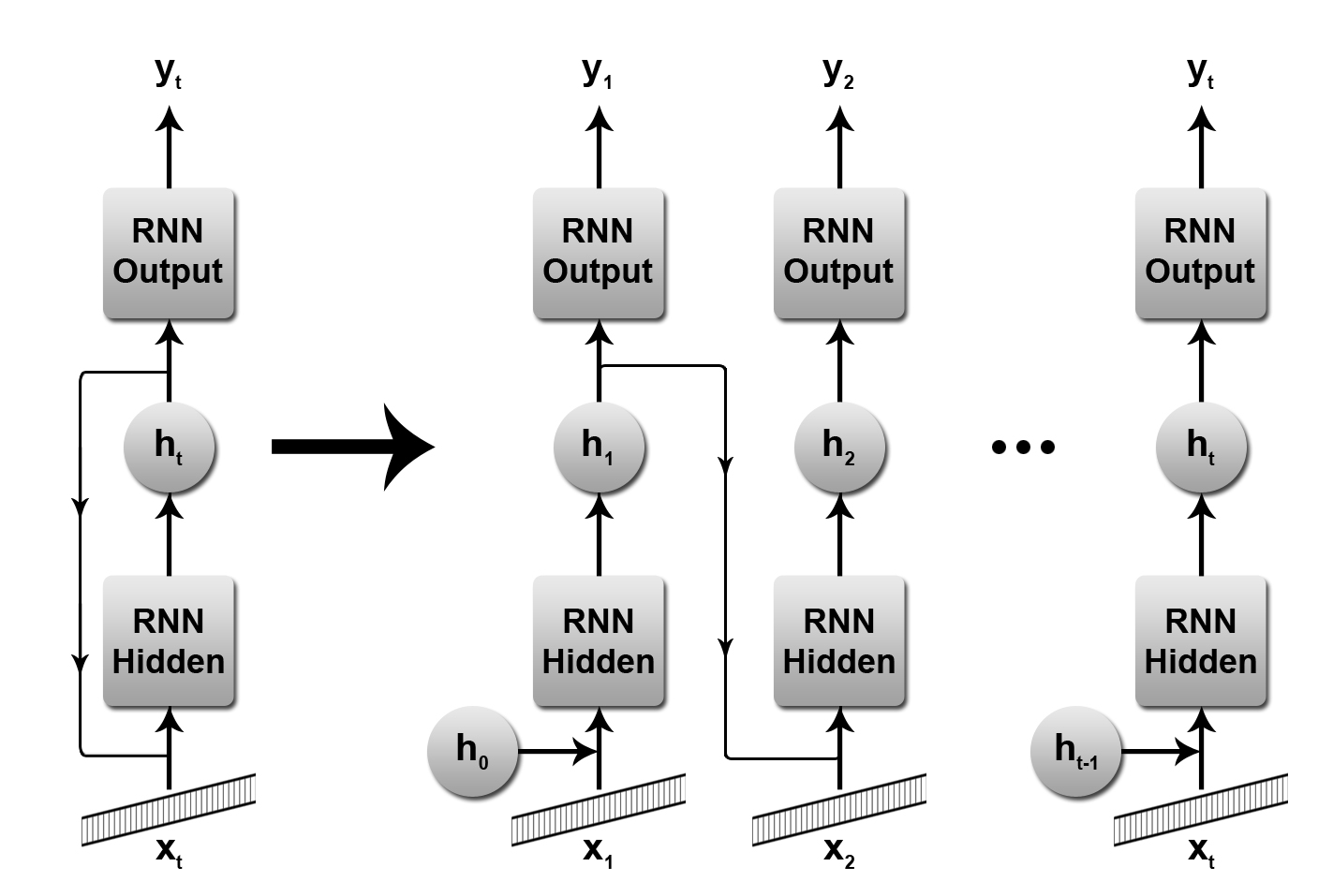}
    \caption{Recurrent Neural Network diagram. The input vector ${\bf x}_t$ is combined with the previous state of the network ${\bf h}_{t-1}$ to get the current state ${\bf h}_t$ and the output $y_t$ at timestep $t$.}
    \label{fig:rnn_diagram}
\end{figure}

Here, the input to the network at each timestep, ${\bf x}_t$, has a shape corresponding to the 32 EEG channels, resulting in an input tensor of shape: $(\text{\textit{batch\_size}}, \text{\textit{time\_steps}}=32, \text{\textit{channels}}=32)$. This means that at each timestep, the network receives a 32-dimensional vector, one from each electrode. As a result, multiple EEG trials are processed in parallel according to the configured batch size. The following equation describes the model:

\begin{equation}
    {\bf h}_t = tanh({\bf W}^{xh} {\bf x}_t + {\bf W}^{hh} {\bf h}_{t-1} + {\bf b}^h), \label{eq:elman}
\end{equation}

\noindent where ${\bf x}_t$ is the input to the recurrent layer at timestep $t$, ${\bf h}_t$ is the activation vector of the recurrent layer at timestep $t$, ${\bf W}^{xh}$ and ${\bf W}^{hh}$ represent the weights associated with the input vector ${\bf x}_t$ and the hidden state ${\bf h}_t$, respectively, and ${\bf b}^h$ is a bias vector. Here, the recurrent connection is introduced by the term ${\bf W}^{hh} {\bf h}_{t-1}$. At every timestep $t$, the hidden layer computes its output, ${\bf h}_t$, as a function of both the current input, ${\bf x}_t$, and the previous hidden state, ${\bf h}_{t-1}$. In this architecture, the hidden state ${\bf h}_t$ is used to calculate the network's output as:

\begin{equation}
    y_t = \sigma({\bf w}^{hy} {\bf h}_t + b^y), \label{eq:output}
\end{equation}

\noindent where ${\bf w}^{hy}$ is a weight vector and $b^y$ is the bias. Note that, since we are facing a binary classification problem, only one neuron is used in the output layer. The standard procedure for training an RNN consists of unfolding it in the temporal dimension, creating several copies of itself, which share the same weight matrices, for different timesteps, and backpropagating the gradients through them. We show the Elman RNN architecture and its corresponding unfolding in Figure \ref{fig:rnn_diagram}.

When classifying P300 events, we get the final prediction of the Elman RNN from the last output after processing the entire $1000~ms$ (32 timesteps) input sequence. We define the prediction probability, $p$, of an event being a P300 as the network's output for the last timestep ($t=32$): 

\begin{equation}
    p = y_{32}. \label{eq:vanilla_p}
\end{equation}

\noindent The network processes each element of the sequence one by one, retaining information from the previous elements in its hidden state. Ultimately, the last output encapsulates the network's prediction or interpretation of the entire sequence.

\subsection{Training Configuration and Validation Strategy}
\label{subsec:experiments}

The dataset has eight subjects and four sessions per subject, as mentioned earlier. To evaluate the models, we performed a K-Fold cross-validation with K = 4, using three of the sessions for training and the remaining one for test. In order to define the training phase of the neural network model, we have set the following hyperparameters: the number of units in the recurrent layer is 50, the batch size is 16, the learning rate is 0.0003, and the Nadam optimizer \cite{nadam} is used. In paradigms such as the P300 oddball task, there is a natural imbalance because only one out of every six stimuli is a target; without correction, the model would tend to predict the majority class (non-target) for all inputs. To address this, we weighted the training data by assigning lower weights to the majority class (non-targets). This approach effectively mitigates bias towards the dominant class. The network is trained to minimize the cross-entropy between the expected targets, $t_{i}$, and the network's predictions, $p_{i}$:

\begin{equation}
    XE(t, p) = - \frac{1}{N} \sum^{N}_{i=1} [t_i log(p_i) + (1 - t_i) log(1 - p_i)].
    \label{cross_entropy}
\end{equation}

\noindent Finally, for evaluation, we measure the performance using the balanced accuracy (BAC) metric \cite{eeg_oliva21,p300_oliva23_icann,p300_oliva_aiai23}:

\begin{equation}
    BAC = \frac{recall + specificity}{2}.
    \label{eq::bac}
\end{equation}

\noindent The BAC is particularly suitable for imbalanced datasets because it prevents misleading results that could arise from the dominance of the majority class.

\subsection{Methodology for Spatio-Temporal Explainability of RNNs}
\label{subsec:methodology}

Explainability in Deep Neural Networks, particularly Recurrent Neural Networks (RNNs), remains an open challenge because these black-box models lack transparency in their behavior. To address this problem, we propose a methodology for analyzing the explainability of trained RNNs. Our approach serves a dual purpose: first, to establish a procedure for extracting the desired explanations, and second, to provide insight into the temporal and spatial behavior of the P300 ERPs. We first analyze the network's internal behavior by visualizing the activation of the hidden neurons. With this, we aim to identify any unique patterns that emerge when the network processes a target stimulus, assuming such patterns will not be present with non-target stimuli. We anticipate retrieving key information from the hidden activity around $300~ms$, which will lead to further analysis and improved results.

Next, assuming that the internal behaviors are recognized, we propose to analyze the network's input and output weights, as these serve as gateways between the black-box model and the external environment. This idea allows to perform a spatial analysis of the input EEG signals by using the model's weights to evaluate the global relevance $R_i$ of each electrode $i$, which we define as:

\begin{equation}
    R_i = \sum^{50}_{j=1} |{\bf W}^{xh}_{ij}|,
    \label{eq:suma_pesos}
\end{equation}

\noindent where ${\bf W}^{xh}_{ij}$ is the weight connecting the $i$-th input to the $j$-th hidden neuron. Since the electrodes are located at different positions on the scalp, we can identify which brain regions contribute the most to the network's decision-making process. This allows us to compare whether the model relies on the expected regions (parietal, occipital, or frontal areas) where the P300 typically shows higher amplitude. Furthermore, because we are defining the relevance of the input electrodes in terms of the network input weights, we show the critical role of incorporating $L_1$ regularization to achieve enhanced explainability. By applying this penalty, the model is encouraged to focus on the relevant electrodes, while the weights of the others tend to diminish or become negligible. This procedure significantly contributes to the model's explainability and spatial analysis. Specifically, $L_1$ regularization modifies the loss function by adding a term that is proportional to the $L_1$ norm of the weight matrix:

\begin{equation}
    L(t, p) = XE(t, p) + \lambda ||{\bf W}^{xh}||_1, 
\end{equation}

\noindent where $XE(t, p)$ is the cross-entropy loss in Equation \ref{cross_entropy} and
$\lambda$ is a hyperparameter controlling the regularization strength, which is set to $0.1$. 

Last, we propose to analyze how the information flows through the network from the input signals to the final decisions in order to perform a spatio-temporal analysis for each individual sequence or window. Following \cite{Shrikumar_2017}, we define the local relevance of electrode $i$ at time $t$ for input sequence $k$ as follows: 

\begin{equation}
    R^{(k)}_{it} = x^{(k)}_{it} \frac{\partial p^{(k)}}{\partial x^{(k)}_{it}},
    \label{eq:derivada}
\end{equation}

\noindent where $x^{(k)}_{it}$ represents the value associated to electrode $i$ at time $t$ for the input window $k$, and $p^{(k)}$ is the network's prediction (estimated probability of P300 event) for the same window.  This equation involves the idea of gradient, which measures how sensitive the output is to small changes in the input or, in other words, the relevance of the input. Considering the input as a time-series, we can visualize not only the electrodes' relevance (spatial analysis), but also the precise moment when these are relevant (temporal analysis). This approach allows us to verify whether the model correctly identifies the temporal window in which the P300 is expected to emerge, thereby confirming that the network's decisions are based on meaningful neural signals and consistent with established neurophysiological patterns of the P300 across both cortical regions and time intervals.

\section{Explainability of Recurrent Neural Networks}
\label{sec:rnns}

The explainability of Recurrent Neural Networks has played a decisive role in our work. The effort to understand the internal mechanisms within the networks has involved some novel approaches, which have led to a continuous improvement of our models \cite{eeg_oliva21,p300_oliva23_icann,p300_oliva_aiai23}. The study of the explainability of the Elman network holds significant value in enhancing our understanding of the model's internal behavior and the underlying mechanisms that facilitate the identification of P300 events. By delving into the explainability aspect, we can gain valuable insights into how the neural network processes information and arrives at its decisions, shedding light on the factors influencing its performance. 

\subsection{Analysis of the Hidden Layer's Activations}
\label{subsec:prm}

\begin{figure}
    \centering
    \includegraphics[width=\linewidth]{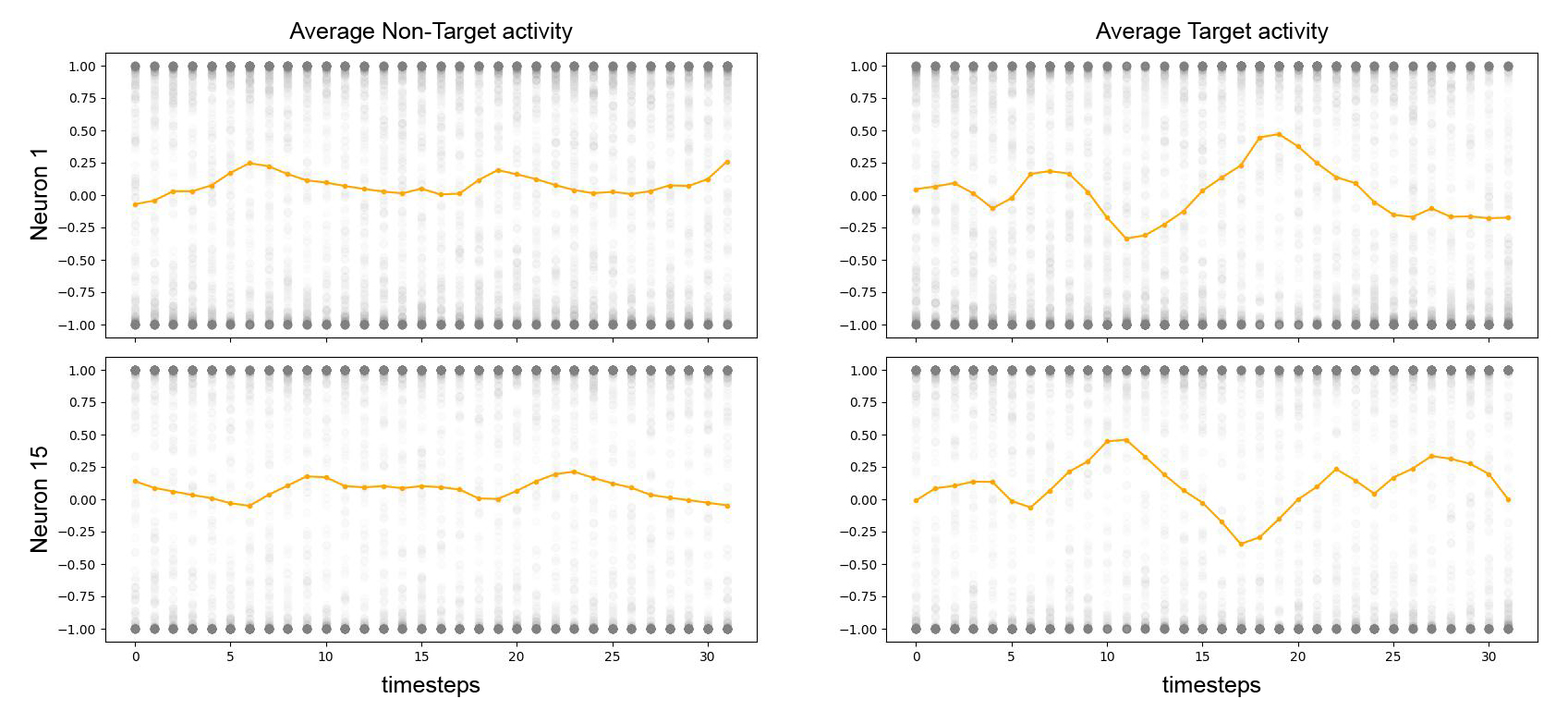}
    \caption{Activity of two selected hidden neurons from an Elman Network trained on user 1, session 1. The left column shows non-target windows, and the right column shows target windows. Gray dots represent the activity for individual windows, and the orange curves correspond to the average activity across all windows.}
    \label{fig:tvsnt_vanilla}
\end{figure}

To understand the internal behavior of the networks, we initiated a comprehensive comparison at the neuron level, carefully analyzing and contrasting the responses evoked when the model encountered a P300 window versus a non-P300 window. This examination allows us to discern distinct patterns and activations within the network and ultimately to use the gained knowledge to improve the model. It is worth noting that, although the model's final prediction is computed solely from the last timestep of the sequence, the recurrent hidden state propagates information from all previous timesteps. Therefore, intermediate inputs are not ignored, but they influence the output because their information is retained and transmitted through the recurrent connections. However, this temporal propagation may degrade or distort over longer sequences, meaning that information from earlier timesteps may have a reduced impact on the final decision. Examples of the activity of two hidden neurons are shown in the different panels of Figure $\ref{fig:tvsnt_vanilla}$. The neuron activity for all non-target (left panels) and target (right panels) windows is plotted with gray dots, and the average activity is superimposed using orange curves. Note that this activity is bounded in the interval $[-1, 1]$ due to the use of the \textit{tanh} activation function in Equation $\ref{eq:elman}$.

\begin{figure}
    \centering
    \includegraphics[width=\linewidth]{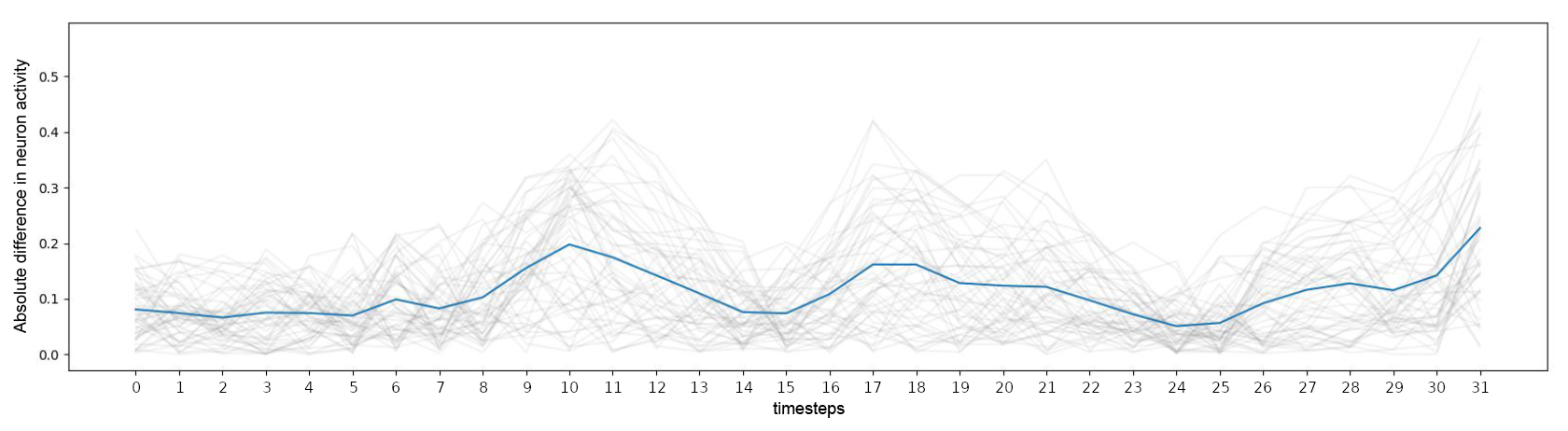}
    \caption{Absolute difference between the average activation for target and non-target windows for each hidden layer neuron at every timestep, in a network trained on user 1, session 1. Each gray curve is for a different neuron, while the blue curve is the average across all neurons.}
    \label{fig:hs_diff_vanilla}
\end{figure}

We observe that the average activity of individual neurons differs significantly when faced with P300 versus non-P300 windows. This dissimilarity is not limited to the final timestep, where decision-making takes place. Instead, neuron activities exhibit numerous variations at earlier intervals, and these fluctuations may also convey essential information. To validate this assumption, we computed the absolute difference between the average activation for target and non-target windows for each hidden layer neuron at every timestep. The result is shown in Figure $\ref{fig:hs_diff_vanilla}$, where the gray curves represent individual neurons and the blue curve is the average across all neurons.

As expected, the maximum disparity in activity occurs at the final timestep, which represents the completion of the decision-making process. However, in this example, we also observe an increased difference in neuron activity during other time intervals, specifically between $250~ms$ and $375~ms$ and again between $500~ms$ and $687.5~ms$. This suggests that relevant information for distinguishing between target and non-target windows appears earlier in the temporal sequence. Crucially, the model does not exploit these intermediate signals since its prediction depends solely on the final hidden state. In other words, while the RNN produces an output only at the last timestep, information from earlier discriminative patterns may be reduced even though it is propagated through the recurrent connections. Therefore, the network could be discarding valuable information.

\begin{figure}
    \centering
    \includegraphics[width=\linewidth]{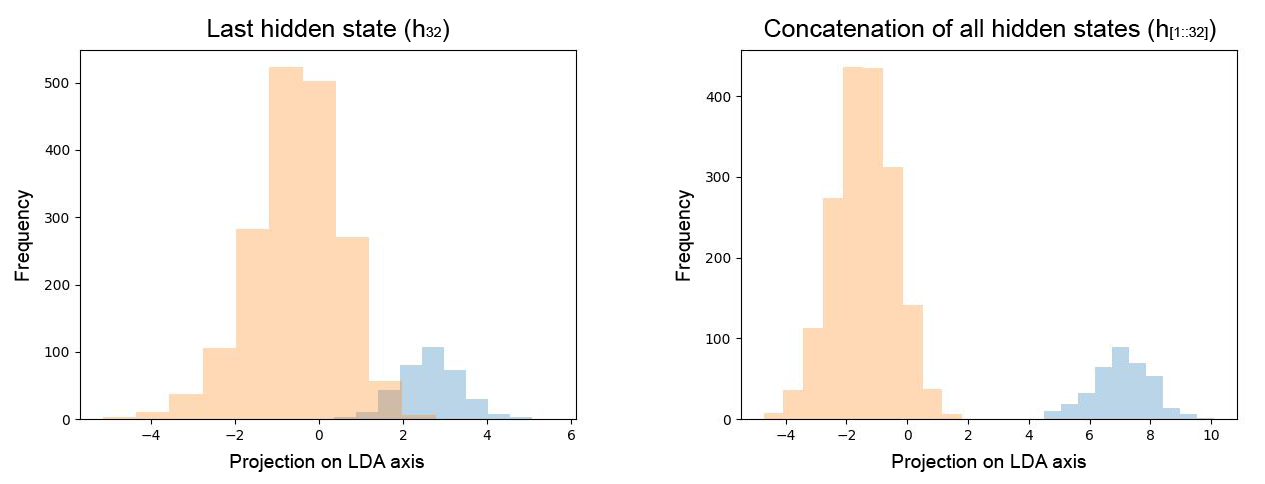}
    \caption{Distribution of target (blue) and non-target (orange) windows projected onto the LDA axis. The left plot uses the final hidden state, $\mathbf{h}_{32}$, as input, and the right plot uses the concatenated hidden states across all timesteps, $\mathbf{h}_{[1:32]}$. Network trained on user 1, session 1.}
    \label{fig:vanilla_lca}
\end{figure}

To further illustrate this limitation, we applied Linear Discriminant Analysis (LDA) to evaluate the separability of the target and non-target classes, using information from either only the last hidden state or the entire set of 32 hidden states. The results are shown in Figure $\ref{fig:vanilla_lca}$, where the left plot represents the LDA projection of all target and non-target windows when only the last hidden state, ${\bf h}_{32}$, is considered. The right plot presents the same analysis using the concatenation of hidden states across all timesteps, ${\bf h}_{[1::32]}$. This comparison highlights the significance of utilizing information from all timesteps instead of only the last one, as the former yields improved discriminative power between the target and non-target classes.
This analysis sheds light on the temporal dynamics of internal representations and reveals how the integration of information over time can enhance classification performance. Consequently, gaining a comprehensive understanding of the contribution of different time points in the classification process with explainability can improve the effectiveness of our model to predict the P300 response. 

\subsection{The Post-Recurrent Module}

It is important to note that, although the Elman RNN has recurrent memory and can transmit information across the entire sequence, the final decision based solely on the last hidden state may cause some relevant information to be diluted or have reduced influence on the output. The analysis of the previous section highlights the need for mechanisms that explicitly capture intermediate temporal contributions. This motivates an alternative approach that considers not only the last timestep but the entire sequence to make the final decision. Specifically, we propose the addition of a new processing layer, the Post-Recurrent Module (PRM) \cite{p300_oliva23_icann}, which processes the temporal information extracted by the recurrent layer across all time steps to obtain the final network's output. This layer operates on the concatenation, $\textbf{y}_{[1::32]}$, of the recurrent layer outputs for each of the 32 time steps:

\begin{equation}
    p = \sigma({\bf w}^p {\bf y}_{[1::32]} + b^p), \label{eq:one-neuron-dense}
\end{equation}

\noindent where ${\bf w}^p$ is a weight vector and $b^p$ is the bias. This approach allows the model to learn how to combine and scale the contribution $y_t$ of each timestep to produce the final output $p$. In other words, the PRM acts as an explicit temporal integration mechanism, preventing the network from relying exclusively on the compressed information in the last hidden state and allowing each timestep to contribute directly to the output. This enables the model to identify intervals of interest for P300 event detection, enhancing both the accuracy and robustness of the decision-making process. Table \ref{tab:results} shows the BAC obtained for both models (with and without PRM). For reference, the result from the BLDA model proposed by Hoffmann et al. \cite{hoffman} is also included. Results are averaged across all users and sessions. 

\begin{table}
    \centering
    \caption{Test Balanced Accuracy (BAC) averaged across all users and sessions with cross-validation K=4. The first column shows the baseline result using BLDA, and the last two columns show the Elman RNN results with and without the PRM layer.}
    \begin{tabular}{lcccc}
        \hline
        Model: & \textit{BLDA} \cite{hoffman} & \textit{Elman} \cite{p300_oliva_aiai23} & \textit{Elman PRM} \cite{p300_oliva23_icann} \\
        Test BAC: & 0.71$\pm$0.05 & 0.71$\pm$0.06 & \textbf{0.80$\pm$0.05} \\
        \hline
    \end{tabular}
    \label{tab:results}
\end{table}

It is worth noting that the Elman RNN enhanced with the PRM layer achieves the best BAC (0.80), improving the standard RNN and the baseline results by 9\%. This improvement confirms that temporal information not fully exploited by the standard Elman RNN is indeed relevant for discriminating P300 from non-P300 events, and that the PRM enables the model to leverage this information more effectively.

\subsection{Interpreting Temporal Contributions via the PRM Weights}

\begin{figure}
    \centering\includegraphics[width=\linewidth]{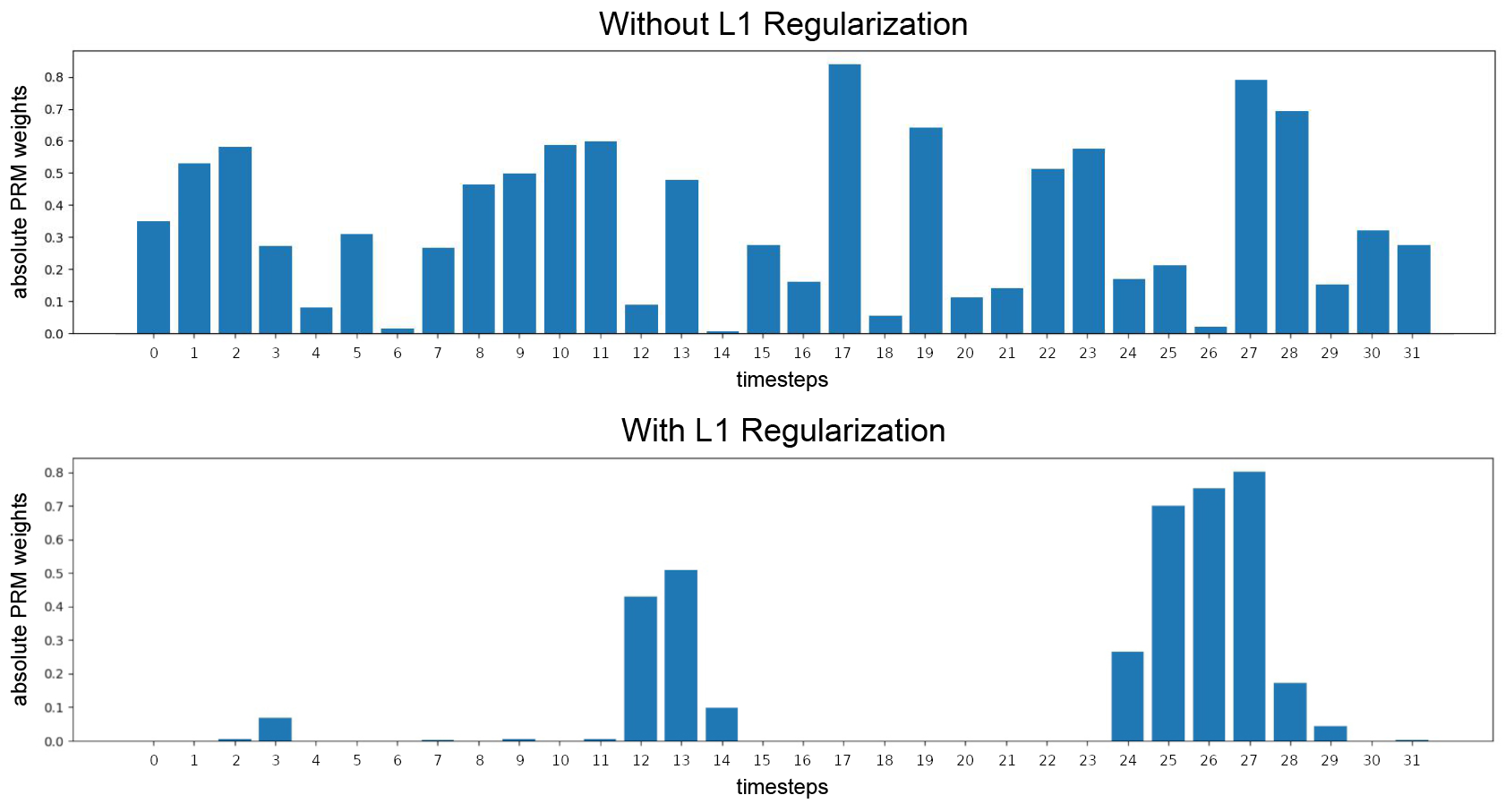}
    \caption{Absolute value of the PRM weights, \textbf{w}$^p$, from two networks trained on user 1, session 1. The network in the top panel was trained with no regularization. The network in the bottom panel was trained with $L_1$ regularization applied to the PRM weights, with regularization parameter $\lambda = 0.01$.}
    \label{fig:lr-weights}
\end{figure}

The PRM strategy combines the 32 RNN outputs from a single $1000~ms$ window through the use of logistic regression, as outlined in Equation \ref{eq:one-neuron-dense}. By delving into this equation, we open up the opportunity to scrutinize the logistic regression weights, \textbf{w}$^p$, to visually discern the intervals of interest. This analysis not only grants a deeper understanding of the neural network's inner workings, but also provides a means to interpret the model's decision-making process. Furthermore, we show the critical role of incorporating regularization in the PRM connections to improve the model performance and to achieve enhanced explainability of the results. Figure \ref{fig:lr-weights} displays the PRM weights (in absolute value) of two networks trained on user 1, session 1. The network in the upper plot was trained with no regularization, while the network in the lower plot included $L_1$ regularization, with $\lambda = 0.01$, in the PRM weights. 

The figure underscores the crucial role of regularization, acting as a filter against superfluous or redundant temporal instances. This mechanism effectively discards unnecessary information, skillfully sidestepping the noise that may otherwise taint the signal at various time intervals. The comparison between the two networks highlights the importance of combining the PRM with $L_1$ regularization to optimize the model's explainability. In particular, the regularized model shows relevant active regions around $375$-$500~ms$ (12-16 timesteps) and $750$-$900~ms$ (24-29 timesteps). The timing of the first active region, around $250$-$500~ms$, aligns with the classic P300/P3b component described by Polich \cite{Polich2007}, which is characterized as a late positive potential typically emerging after target stimulus presentation in oddball paradigms. This confirms that the network effectively identifies the canonical P300 response. The meaning of the second active region, approximately between $750$-$900~ms$ is not as clear and admits several interpretations.

On the one hand, from a pure neuroscientific perspective, the network might be identifying other types of event, such as N750 and N900, reported by Lytaev \cite{lytaev2021} in visual oddball paradigms, which are associated with late-stage visual processing, cognitive integration, and stimulus discrimination. On the other hand, from a more genuine machine learning point of view, the second peak could be an artifact related to the overlap between the time windows characterizing the EEG signals for two consecutive images (see Section \ref{subsec:dataset}). Admitting the second possibility, the increased PRM weights around $750~ms$ could be reflecting the necessity to detect a second potential P300-like response associated with the next stimulus. This detection could be an indicator that the current image does not correspond to a target. 

\begin{figure}
    \centering
    \includegraphics[width=\linewidth]{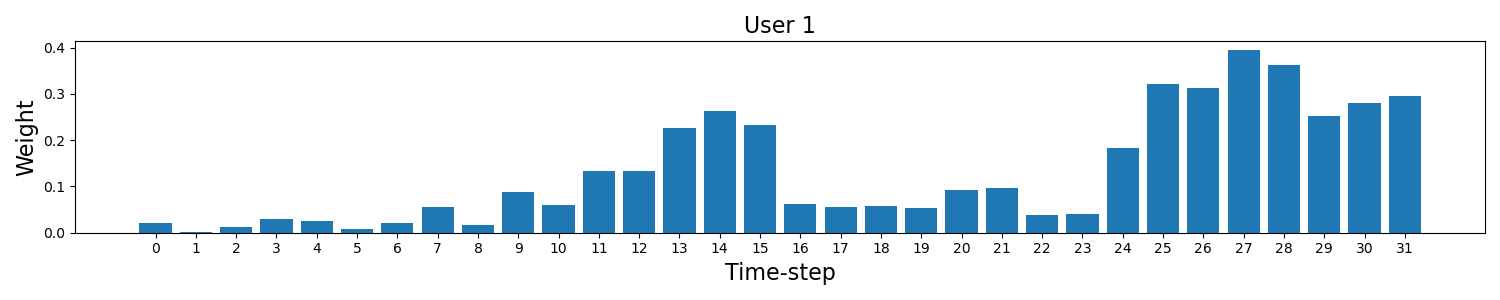}
    \includegraphics[width=\linewidth]{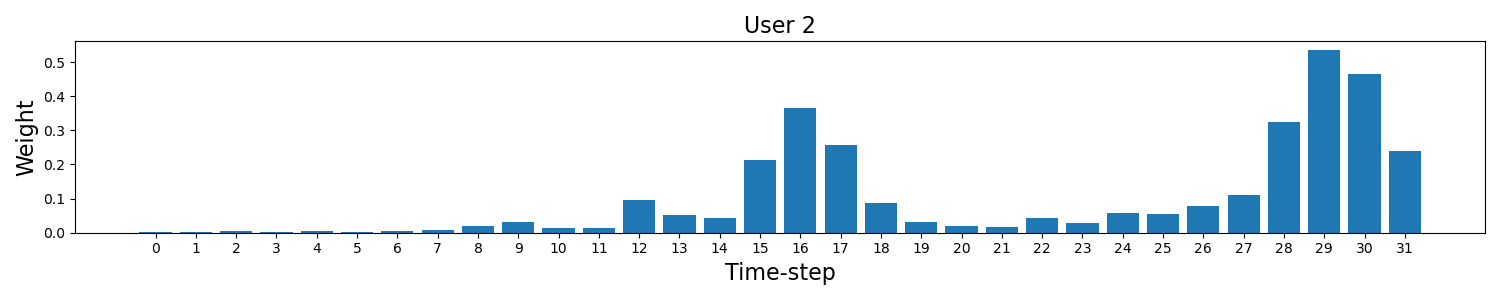}
    \includegraphics[width=\linewidth]{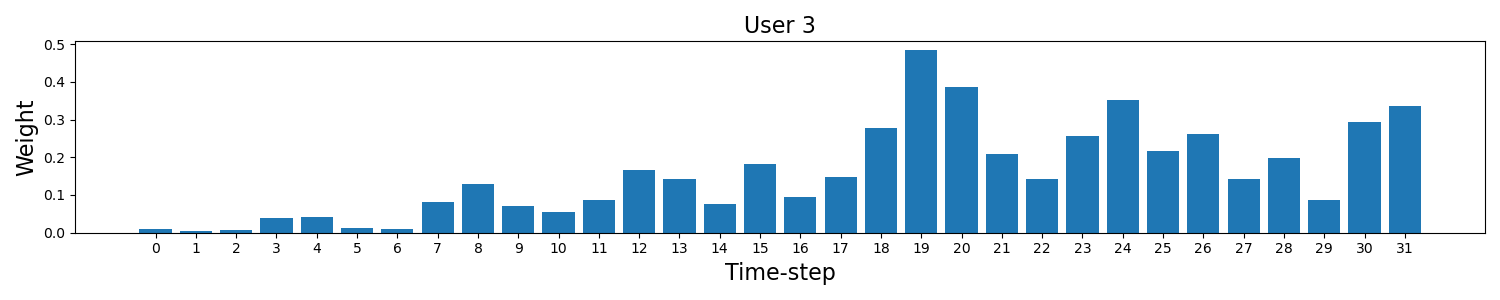}
    \includegraphics[width=\linewidth]{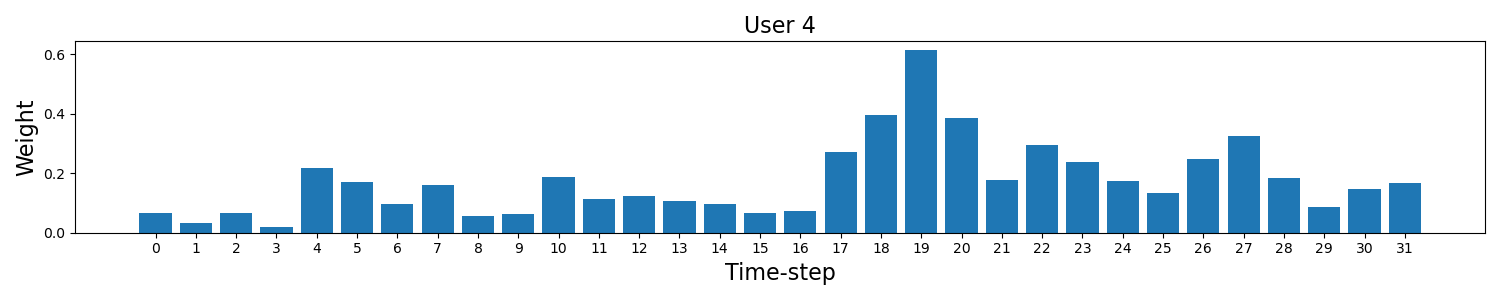}
    \includegraphics[width=\linewidth]{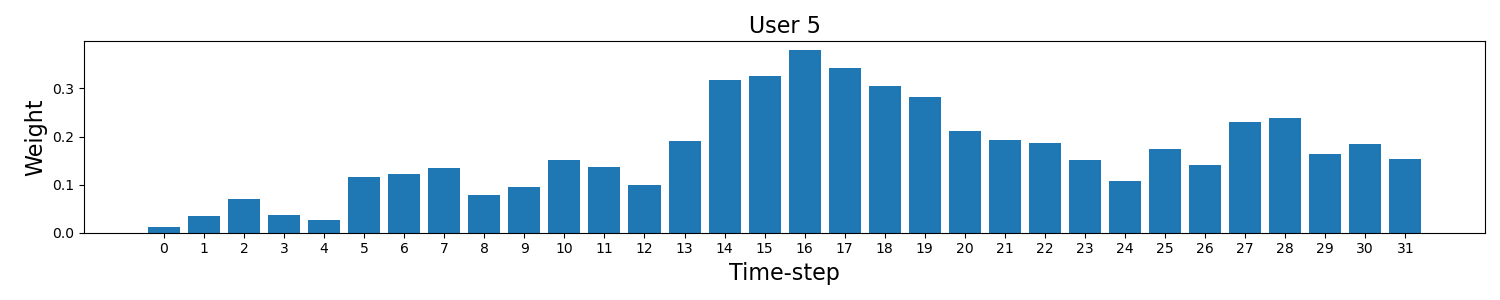}
    \includegraphics[width=\linewidth]{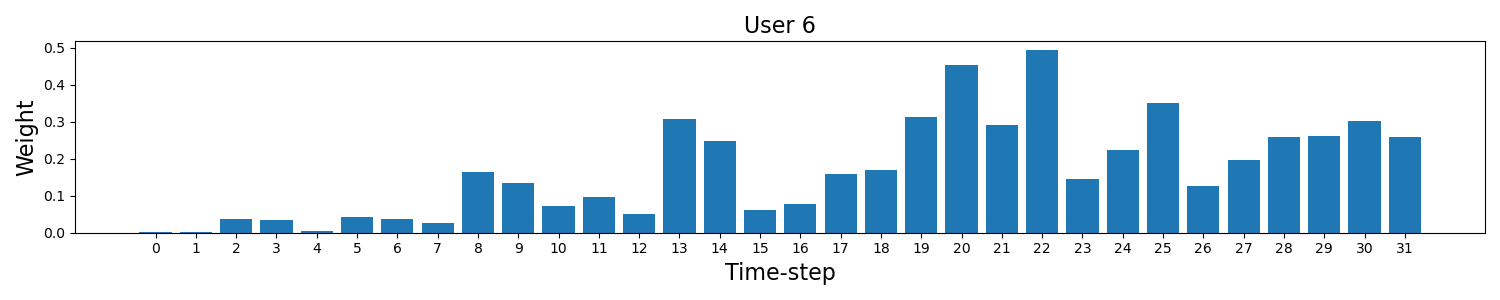}
    \includegraphics[width=\linewidth]{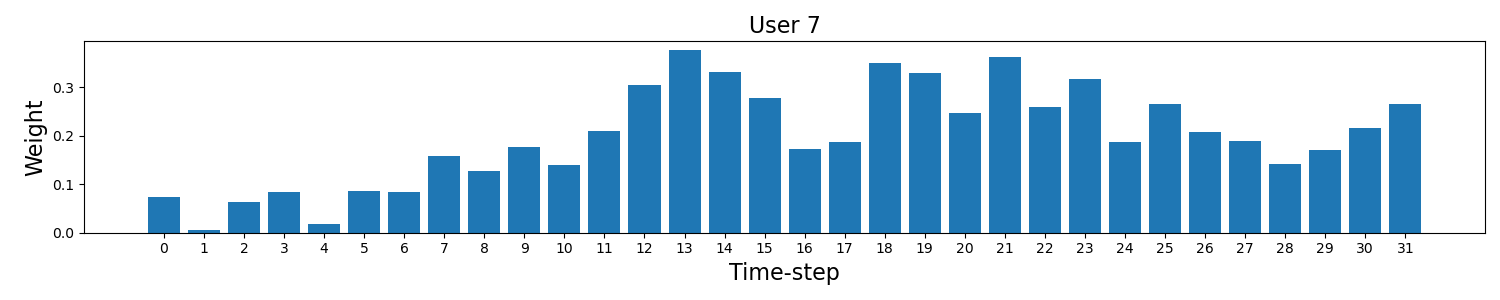}
    \includegraphics[width=\linewidth]{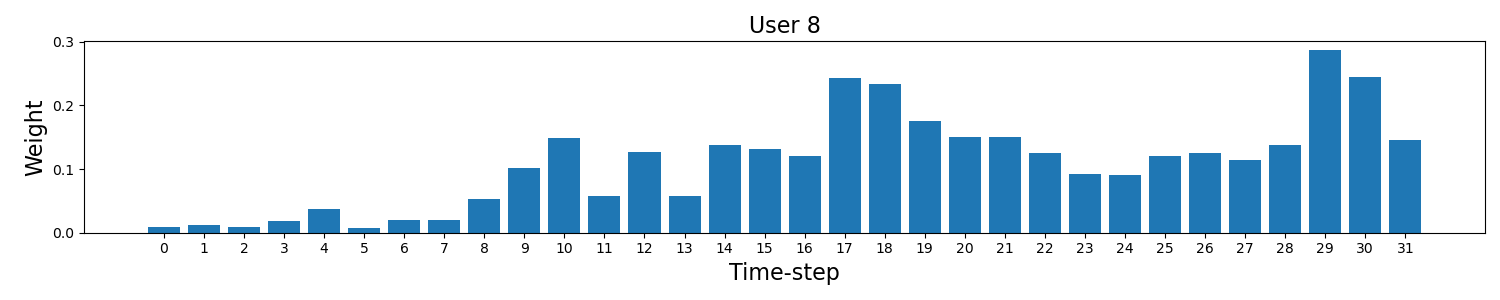}
    \caption{Absolute value of the PRM weights averaged across 400 networks (100 per session) trained on users 1 to 8. All networks included $L_1$ regularizarion in the PRM weights, with $\lambda = 0.01$.}
    \label{fig:lr-weights-avg}
\end{figure}

In order to test the generality of the previous result, we repeated the experiment, training 400 different networks (100 per session) on every user. The absolute values of the PRM weights, averaged across networks, are shown in Figure \ref{fig:lr-weights-avg} for each user. These weights suggest that the neural networks adeptly harness all the relevant information starting from $300~ms$ onward, including the second region after $750~ms$. It is important to note that, since the model can analyze broad temporal windows, it captures not only the classic P300 peak but also subsequent dynamics that may be related to stimulus interpretation. This is in line with several studies, particularly \cite{qin2022}, which argue that brain activity relevant to the P300 event is not limited to the $300~ms$, but extends to later phases. The differences between users not only align with the previous results, but also highlight a significant inter-subject variability. In summary, the study of the PRM weights presents compelling evidence of the evolving ability of networks to differentiate stimuli and their remarkable adaptability to leverage temporal dynamics for accurate predictions, with 0.80$\pm$0.05 balanced accuracy on average for the eight available subjects. This reinforces the idea that recurrent networks can capture broader cognitive dynamics than classic P300 detectors, providing a richer perspective on brain processing.

\subsection{Spatial Analysis of EEG Inputs for P300 Localization}
\label{subsec:spatial_analysis}

\begin{figure}
    \centering
    \includegraphics[width=\linewidth]{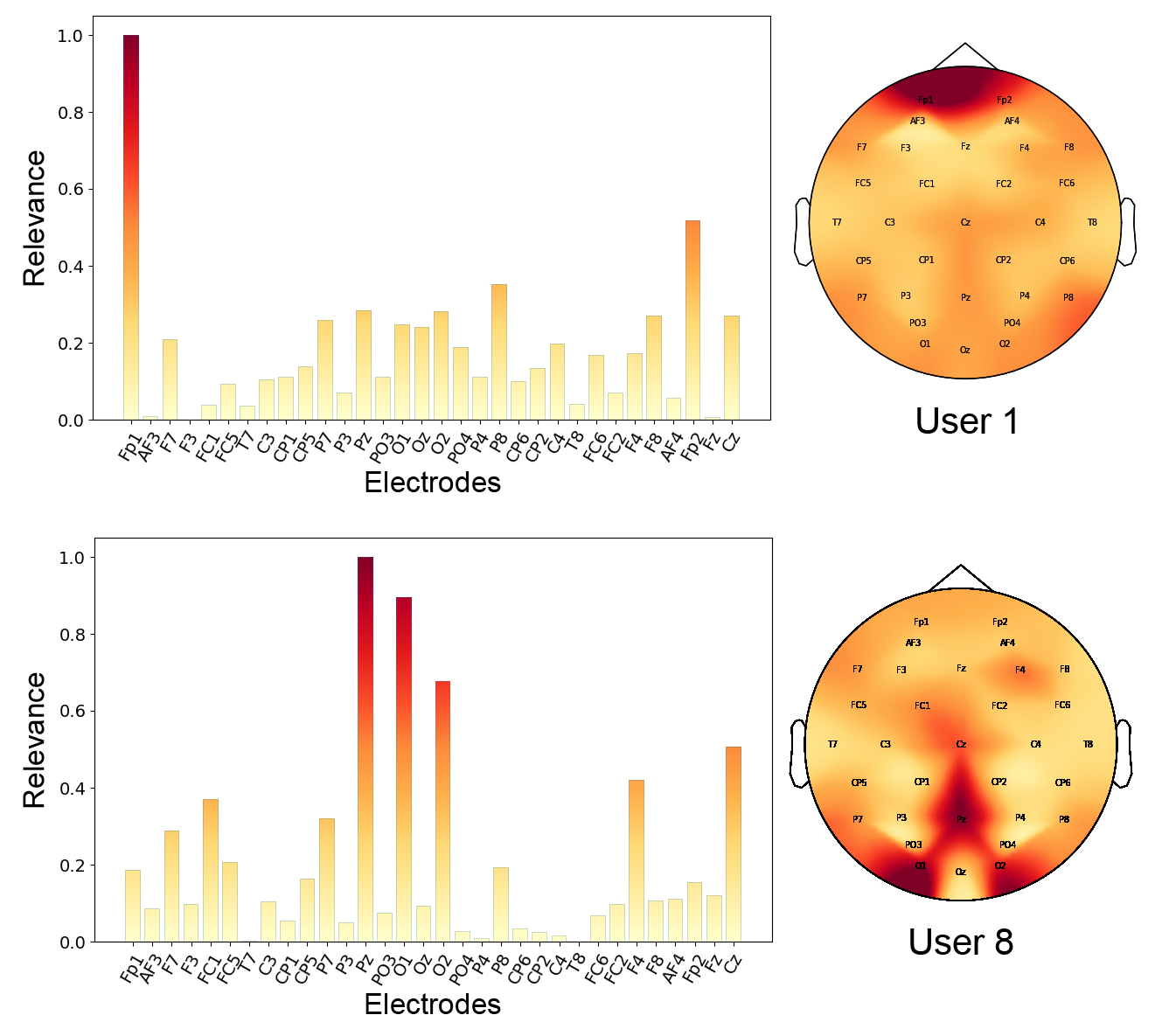}
    \caption{Top: figures related to user 1. Bottom: figures related to user 8.
    Left: Bar diagram of the input layer relevances (see Equation \ref{eq:suma_pesos}) from a network trained on user 1 (top) or 8 (bottom) and test session 1 in both samples. Vertical axis represents the normalized relevances, being 1 the most relevant and 0 the least one. Right: Head plot of the same relevances.}
    \label{fig:wxh-user0}
\end{figure}

Following in the context of identification of P300 ERPs, while our temporal analysis unravels the precise timing of P300 responses, the counterpart lies in spatial analysis, allowing for the localization of brain activity responsible for these ERPs. Integrating both dimensions enables us to verify whether the model reproduces known neurophysiological patterns, generally centered in the parietal region. This section delves into the importance of spatial analysis and shows how it contributes to the identification and interpretation of P300 responses. First of all, we analyze the relevance of the input layer as described in Equation \ref{eq:suma_pesos}. This proposal allows us to analyze the electrodes that convey the relevant information that the network must retain. In Figure \ref{fig:wxh-user0} we show two head plots, representing the relevance $R_i$ of the input layer of two networks trained on users 1 and 8. In both figures, the left plot represents a bar diagram with relevance $R_i$, and the right one is its equivalent topographical head plot. 

We can observe in the first example (user 1, Figure \ref{fig:wxh-user0}-top) that the most relevant information occurs in the frontal lobe ($Fp1$), followed by $Fp2$ and parietal $P8$. The rest of the electrodes also contribute, but not on the same scale. On the other hand, in the second example (user 8, Figure \ref{fig:wxh-user0}-bottom), we observe that this time the most relevant information occurs in the occipital ($O1$, $O2$) and parietal ($Pz$) lobes, and again, the rest of the electrodes contribute on a lower scale. With these two examples, we show again the importance of considering the inter-variability between subjects since it does not only appear at different time windows but also in different physical lobes in the brain. 

This spatial variability is consistent with previous studies that have reported substantial differences in P300 topography both across subjects and experimental conditions. The P300 peak can be maximal in parietal regions during bottom-up attention tasks, but shift toward frontal areas in top-down tasks \cite{Zhang2021}. It can also show prominence in the mid-parietal region, particularly at electrode $Pz$ \cite{Li2019}, and parieto-occipital distributions during visual oddball tasks, highlighting the relevance of electrodes such as $O1$ and $O2$ \cite{Ponomarev2025}. These findings support the patterns observed in our users and reinforce the need to account for inter-subject variability when interpreting P300-based neural network models.

\subsection{Spatio-temporal Explainability of RNNs in P300 Detection}
\label{subsec:spatiotemporal_analysis}

\begin{figure}
    \centering
    \includegraphics[width=\linewidth]{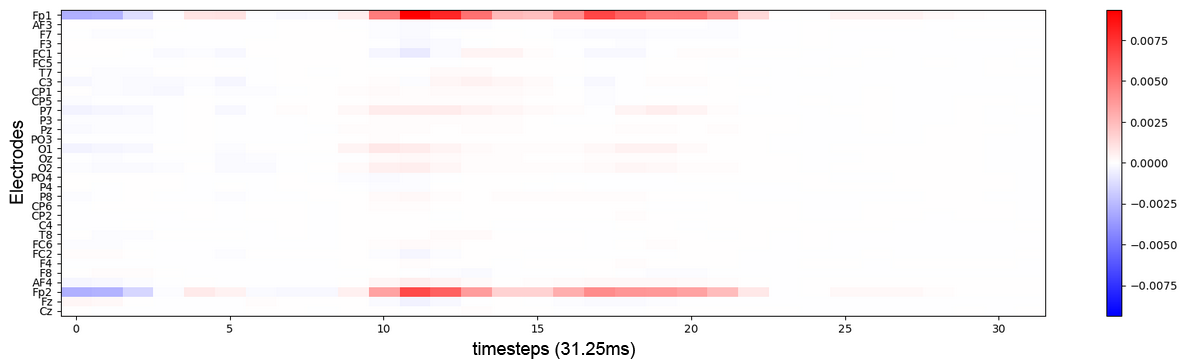}
    \includegraphics[width=\linewidth]{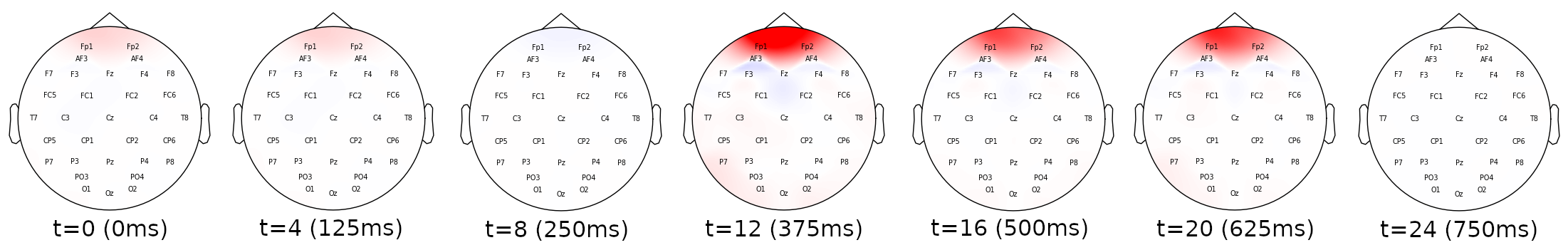}
    \caption{Spatio-temporal analysis of the P300 signal extracted from a network trained on user 1 when facing a target stimulus. The top figure represents the relevance $R_i$ extracted when the network is trained with regularization, where the horizontal axis represents the timesteps, and the vertical axis represents the electrodes. The bottom figure represents the same $R_i$ in a topographical head plot. In both figures, the color represents the relevance itself. Red color is interpreted as high relevance for classification, and blue color is interpreted as inverse relevance (or negative) for classification, which is interpreted as relevant information for missclassification.}
    \label{fig:spatiotemporal-deriv}
\end{figure}

Integrating both temporal and spatial analysis allows us to gain valuable insights into the spatio-temporal organization of the P300 response, and at the same time, enhances the explainability of the recurrent neural networks. To do so, we distinguish the spatio-temporal analysis performed when identifying non-target stimuli from that used when identifying the target stimulus. Then, our explainability is based on the visualization of the relevance defined in Equation \ref{eq:derivada} for both target and non-target situations when the networks are trained with and without L1 regularization. We show in Figure \ref{fig:spatiotemporal-deriv} the average target spatio-temporal relevance for user 1 with $\lambda = 0.1$ input regularization. In the upper figure, timesteps are represented in the horizontal axis, and each of the electrodes is identified in the vertical axis. In the lower figure, the same information is represented in topographical head plots. It is important to highlight the distinction between positive and negative relevance. The red color represents high relevance (positive), which is interpreted as relevant information for classification, and indicates parts of the signal that support correct classification of the P300. The blue color represents inverse relevance (negative), highlighting patterns that could lead the model to misclassify. 

We can obtain information from these figures. First, the crucial role of L1 regularization is again highlighted. The network extracts the most relevant information from the input data, enhancing not only the accuracy (as shown in the previous Section \ref{subsec:prm}) but also its explainability. We can observe a direct connection with the previous spatial analysis (see Figure \ref{fig:wxh-user0}) since the most relevant information remains in the same electrodes ($Fp1$ and $Fp2$). However, with this analysis, we also deepen the temporal analysis and observe how the relevant information is mainly in the first 20 timesteps (approximately the first $625~ms$). In addition, from the shape of the relevance, we can confirm that something happens in the $Fp1$ and $Fp2$ electrodes when facing the target stimulus. These results show how the information flows in the brain between the electrodes and how the network identifies these spatio-temporal dependencies, and also verify that the network's decisions are not arbitrary, but are based on plausible neural patterns related to the dynamics of the P300.

\section{Discussion and Conclusion}
\label{sec:conclusion}

In this work, we have proposed an interpretable Recurrent Neural Network (RNN) for classifying P300 signals, with a particular focus on understanding both spatial and temporal aspects of the model's decision-making process. Starting from a standard RNN, we introduced the Post-Recurrent Module (PRM), which not only improved classification performance but also enabled the temporal analysis of activations at the output layer. By combining the PRM layer with the L1 regularization penalty, we achieved a dual explainability perspective: identifying the most relevant electrodes (spatial analysis) and determining when this information becomes crucial for classification (temporal analysis). In addition, the gradient-based local explainability (see Equation \ref{eq:derivada}) allows us to perform a spatio-temporal analysis, which reinforces the previous two and is aligned with the neuroscience state of the art, which argues that a P300 ERP occurs at approximately $300~ms$ onward with a certain inter- and intra-subject variability (temporal analysis) in the parietal and occipital brain regions, since these areas participate in the attention mechanisms \cite{kutas_erp_2011,Luck2014,Polich2007}, but also in the frontal cortex \cite{Blankertz2011,Luck2014} (spatial analysis). These neurophysiologically coherent patterns extracted by the recurrent model support the plausibility of its predictions. Thus, this explainability approach enhances the transparency of the model's behavior and supports the design of more robust and adaptive systems. Moreover, explainability also facilitates subject-specific adaptation and contributes to bias reduction. Overall, these results are consistent with the motivation stated in the Introduction, where the need for models that not only achieve high classification performance but also remain interpretable and aligned with established neurophysiological knowledge was emphasized.

An interesting conclusion from our analysis is that even a standard RNN can detect early markers of target stimuli within the EEG sequence, although it does not use them effectively without the PRM layer. This additional layer captures these markers and processes them properly, achieving an improvement of 9\% in balanced accuracy over the state of the art and demonstrating the effectiveness of our approach. Furthermore, the results reinforce the significance of inter- and intra-subject variability in P300 signal processing, highlighting the need for adaptable systems capable of handling such fluctuations. In summary, the proposed method is conceptually simple yet highly effective. By enabling spatio-temporal explainability, our approach reaches a meaningful impact on the current state of P300 ERP recognition. Specifically, identifying both the key electrodes and the relevant time intervals allows for a substantial reduction in computational cost, which is particularly important in healthcare settings, where explainability is critical. This paves the way for the development of low-cost, embedded BCI systems optimized for real-time applications, which is an essential step for the practical deployment of more transparent and scalable P300-based technologies. This argument is consistent with the motivation of the paper, which emphasizes the need for explainable and robust models for real-world and healthcare applications.

This study opens up several promising directions for further research. First, we can generalize the explainability to other tasks based on EEGs. The explainability framework developed in this work could be extended to other EEG classification problems beyond P300 detection, such as motor imagery, SSVEPs, or cognitive workload assessment. Exploring the applicability of RNNs and the PRM layer in these new contexts may reveal broader patterns in brain activity and improve system generalization. Second, real-world deployment on low-cost embedded systems. Identification of key spatial and temporal features offers an opportunity to simplify the models and reduce computational requirements, which is essential for portable or clinical applications where resources are limited. And third, building on the previous point, we can deepen the explainability of recurrent networks. Although our approach improves transparency, more research is needed to better understand the internal mechanisms of recurrent networks at neural-, layer-, or global-level. Exploring additional explainability techniques, such as attention mechanisms, could lead to even more transparent and trustworthy models.

\section*{Acknowledgements}

This work was supported by PID2023-149669NBI00 (MCIN/AEI and ERDF – “A way of making Europe”), and by Universidad Politécnica Salesiana Project 123-03-2024-06-10.

\bibliographystyle{cas-model2-names}
\bibliography{mybib}

\end{document}